\documentclass[journal]{IEEEtran}

\ifCLASSINFOpdf
\else
   \usepackage[dvips]{graphicx}
\fi
\usepackage{url}

\usepackage{graphicx}
\usepackage{lmodern}
\usepackage{array}
\usepackage{longtable}
\usepackage{tikz}
\usepackage{pgfplots}
\pgfplotsset{compat=1.18}
\usepackage{subcaption}
\usepackage{tipa}
\usepackage{tkz-graph}
\usepackage{tipa}
\usepackage{pgfplots}
\usepackage{booktabs}   
\usepackage{amsmath}
\pgfplotsset{compat=1.18}
\usepackage{caption}
\usepackage{subcaption}
\usepackage{booktabs}
\usepackage{pgfplots}
\usepackage{pgfplotstable}
\pgfplotsset{compat=1.18}
\usepackage{lmodern}
\usepackage{array}
\usepackage{longtable}
\usepackage{tikz}
\usepackage{pgfplots}
\pgfplotsset{compat=1.18}
\usepackage{subcaption}
\usepackage{tipa}
\usepackage{multirow}
\usepackage{booktabs} 
\usepackage{amsthm, amssymb}
\usepackage{adjustbox}
\usepackage[most]{tcolorbox}
\usepackage{xcolor}
\definecolor{mypink}{HTML}{BFBFFF}

\begin{document}

\title{Enhancing speech representation learning with cross-modal knowledge transfer with HGNN under low resource settings: the case study of Yemba}

\author{Yannick Yomie Nzeuhang, Marie Tahon, and Paulin Melatagia Yonta, 
\thanks{This work has been funded by the European Union’s Horizon 2020 research and
innovation program under the Marie Skłodowska-Curie grant agreement No 101007666. The authors tkank the LIUM for its computing resources. Submission date for review  /03/2026.}
\thanks{Yannick Yomie Nzeuhang  is with Department of computer sciences of University of Yaounde I, Yaounde, 812, Cameroun (e-mail: yynzeuhang@gmail.com).}
\thanks{Marie Tahon,  is  with LIUM, Le Mans Université, Av. Olivier Messiaen, 72085 Le Mans, France (e-mail: marie.tahon@univ-lemans.fr).}

\thanks{Paulin Melatagia Yonta is with Department of computer sciences of University of Yaounde I and IRD, UMMISCO, Bondy, France, F-93143(e-mail: paulinyonta@gmail.com) }}

\markboth{Journal of \LaTeX\ Class Files, Vol. 14, No. 8, August 2015}
{Shell \MakeLowercase{\textit{et al.}}: Bare Demo of IEEEtran.cls for IEEE Journals}
\maketitle

\begin{abstract}
Acoustic representation learning is crucial for speech processing, yet low-resource languages (LRLs) face severe data scarcity, limiting the effectiveness of traditional and self-supervised methods. As a promising alternative, in this work, we propose to enhance acoustic representation trough a cross-modal transfer knowledge approach, based on heterogeneous graph neural networks (HGNNs), where acoustic and linguistic entities are modeled as distinct node types within a unified graph. Through message-passing mechanisms, linguistic nodes explicitly transfer knowledge to acoustic nodes, enabling structured and interpretable cross-modal information flow. To highlight this knowledge transfer and its benefits, we measured standard clustering metrics as an intrinsic evaluation of acoustic representation, and to emphasize applicability, we performed isolated-word recognition tasks using an English benchmark and a Cameroonian language dataset in low resources settings . Results demonstrate that acoustic representations consistently benefit from linguistic knowledge propagated through the graph. To our knowledge, this is the first demonstration of explicit cross-modal knowledge transfer for acoustic representation learning using HGNNs, highlighting a promising direction for speech representation in low-resource settings.
\end{abstract}


\IEEEpeerreviewmaketitle

\section{Introduction}

\IEEEPARstart{A}coustic representations play a central role in speech processing, as they transform raw speech signals into numerical embeddings used for downstream tasks such as keyword spotting or speech recognition. While traditional handcrafted features and self-supervised learning (SSL) approaches have achieved strong performance~\cite{wav2vec,wavLM,hubert}, these methods primarily rely on acoustic information alone and typically require large amounts of data to generalize effectively. This limitation is particularly critical for low-resource languages (LRLs), where audio data scarcity hinders the learning of robust acoustic representations~\cite{magueresse}.
Recent multilingual and cross-modal --namely text-speech-- SSL models attempt to alleviate this issue by transferring knowledge from high-resource languages through large-scale pre-training~\cite{xlsr,xlsr-53,SONAR, SAMU-XLSR}. However, this transfer remains implicit, occurring at model parameters level, aside the fact that they require important amount of data. This paper investigates whether linguistic knowledge can be explicitly transferred to acoustic representations, and whether such a transfer can be beneficial in low-resource settings. To this end, we propose a cross-modal approach based on heterogeneous graph neural networks (HGNNs)\cite{gnn_1}, where acoustic and linguistic modalities are modelled as distinct node types within a unified graph. Thanks to message-passing mechanisms~\cite{MPNN} natively applied in GNNs,  linguistic nodes directly propagate information to acoustic nodes. This propagation is the key point to enhance acoustic representation which the purpose of this work. 
Recent applications of graph neural networks (GNNs)~\cite{gnn_1} in audio processing, including emotion recognition~\cite{graphMFT, Shirian, liu_emotion}, speaker diarization~\cite{dia1,dia_2}, and audio classification~\cite{Castro_audio,Shirian,Shilei_few_shot}, demonstrated their effectiveness and versatility for learning audio representations with reasonable amount of data. The core innovation of our work lies in constructing heterogeneous graphs containing both acoustic and linguistic nodes, where message-passing algorithms~\cite{MPNN} facilitate information exchange between these complementary modalities to learn enhanced acoustic representations.
Due to the scarcity of annotated data in LRLs in general, and more specifically in Cameroonian language Yemba in particular, the design of lightweight approach at the sentence level is still highly challenging. 
Therefore, adding to  standard clustering metrics measures, the benefit of knowledge transfer is evaluated on the recognition of isolated words as an applicative task. Experiments conducted in low resource settings on English benchmark and Cameroonian language datasets show that acoustic representations consistently benefit from linguistic knowledge transfer through the graph.




\section{Proposed Method}
\label{sec:method}

Our approach relies on two core components: a heterogeneous graph structure and a graph neural network (GNN). 
We build a heterogeneous graph where nodes represent spoken words and their textual transcriptions 
and learn final word representations via a GNN depicted in Fig.\ref{fig:gcn_architecture}.


\subsection{Heterogeneous Graph Construction}

Since speech data lacks an inherent graph structure, we construct a heterogeneous graph 
$\mathcal{G} = (V, E)$ composed of three subgraphs: 
(i) the \textit{acoustic} subgraph, modeling relations between spoken instances, 
(ii) the \textit{linguistic} subgraph, capturing semantic and phonetic relations between transcriptions, and 
(iii) the \textit{acoustic–linguistic} subgraph, linking the two modalities.

\subsubsection{The acoustic sub-graph}
Given a collection of \( N \)  audio samples for training, we construct an undirected weighted graph \( G_{a} = (V_{a}, E_{a}) \) to capture the relationships among the samples, where \( E_{a} \) is the set of all edges between the connected nodes, and \( V_{a} \) the set of nodes. Let $v_{a_{i}}$ and $v_{a_{j}} \in V_{a}$, $SIM_{a}(v_{a_{i}},v_{a_{j}})$ is defined as a similarity measure between their acoustic representations. The acoustic representation can be any existing type of representation.
However, in this paper, we use MFCC; this choice was motivated by the desire to avoid potential biases toward well-resourced languages that are often embedded in pre-trained models. Since our work focuses on a low-resource language, relying on hand-crafted acoustic features allows for a more language-agnostic representation. In addition, MFCCs require significantly lower computational resources, making them more suitable for lightweight systems and resource-constrained environments.
Let  $T(v_{a_{i}})$ denote the set of nodes $v_{a_{j}} \in V_{a}$ whose transcriptions are the same as $v_{a_{i}}$. One hyperparameter is introduced,  $k_{in} \in \mathbb{N^{*}}$ which specifies the number of neighbors to select for a node $v_{a_{i}}$ in $T(v_{a_{i}})$, and $\delta_{a} \in \mathbb{R}^{+}$ a threshold that restricts the number of edges in $G_{a}$. $E_{a}$ is defined as follow: $\forall v_{a_{i}} \in V_{a}$,   $k_{in}$ nearest neighbors are selected from $T(v_{a_{i}})$, an edge $\{v_{a_i}, v_{a_{j}}\}$ is added if $SIM_{a}(v_{a_{i}},v_{a_{j}})> \delta_{a}$, and the weight is set to 1.

 \subsubsection{The linguistic sub-graph}
Given a collection of \( M \) transcribed words, we construct an undirected weighted graph \( G_{l} = (V_{l}, E_{l}) \) as follows: Let $v_{l_{i}}$ and $v_{l_{j}} \in V_{l}$, $SIM_{l}(v_{l_{i}},v_{l_{j}})$ is defined as a similarity measure between their  embeddings. $\delta_{l} \in \mathbb{R^{+}}$, a hyperparameter used as a threshold to restrict the number of edges in $G_{l}$. To construct $G_{l}$, we embed the \( M \) words, then for each  node $v_{l_{i}}  \in V_{l}$ an edge \(\{v_{l_{i}},v_{l_{j}} \}\) is added  if $SIM_{l}(v_{l_{i}},v_{l_{j}})> \delta_{l}$, the weight is set to $SIM_{l}(v_{l_{i}},v_{l_{j}})$.

To get word embeddings, we introduce a method named ``\textit{bag of phonemes}'' which works exactly like a classic ``bag of words'' except that the vocabulary contains phonemes instead of words and we represent words instead of documents. To build the vocabulary we transcribe the \( M \) words in their phonetic form and consider the set of phonemes that appear as a vocabulary. The resulting word embedding is the vector of occurrence of phonemes in the pronounced word. 

\subsubsection{The acoustic-linguistic sub-graph}
 \label{subsec:hetero}
Given a collection of \( N \) audio speech samples for training and a corresponding collection of \( M \) transcribed words, we define the undirected acoustic-linguistic sub-graph  $G_{al}=(V_{al},E_{al})$ to link the nodes of the two preceding sub-graphs, where $V_{al}=V_{a}\cup V_{l}$; $E_{al}$ is the set of edges between nodes of $G_{a}$ and nodes of $G_{l}$. 
To build $E_{al}$, we pre-train an acoustic recognition model ($AM$) (see implementation details in sec.~\ref{sec:implementation}) on a dataset and use it to deduce the probability distributions $prob(v_{a_{i}}=v_{l_{j}})$ that $v_{l_{j}}$ is the transcription of $v_{a_{i}}$. Then, an edge \( \{v_{a_{i}},v_{l_{j}}\} \in E_{al} \) is added with a weight  $AM(v_{a_{i}}=v_{l_{j}})$ if $prob(v_{a_{i}}=v_{l_{j}})>\delta_{al}$.

The complexity of the heterogeneous graph is controlled by limiting each word
transcription to $k_{b} \in \mathbb{N}^{*}$ edges.

 \subsubsection{Graph Convolutional Neural Model}
\label{subsubsec:model}

We employ a Graph Convolutional Network (GCN) of the GraphSAGE type~\cite{GraphSAGE}, which introduces inductive capabilities allowing representation inference for unseen nodes during training. 
Among its four message-passing options, we adopt the \textit{mean aggregator}, whose update rule is ($h^{k}_{u}$ is an embedding of node $u$, and $\mathcal{N}(u)$ a set of neighbors of $u$):
\begin{equation}\label{eq:graphsage}
h_u^{k+1} = \sigma \left( W_{k} \cdot \text{MEAN} \left( \left\{ h_u^{k} \right\} \cup \left\{ h_v^{k} \mid \forall v \in \mathcal{N}(u) \right\} \right) \right)
\end{equation}

\subsubsection{Objective function }
  \label{subsec:loss}
\begin{figure*}[t]
    \centering
    \begin{subfigure}{0.46\textwidth}
        \centering
        \includegraphics[width=\linewidth]{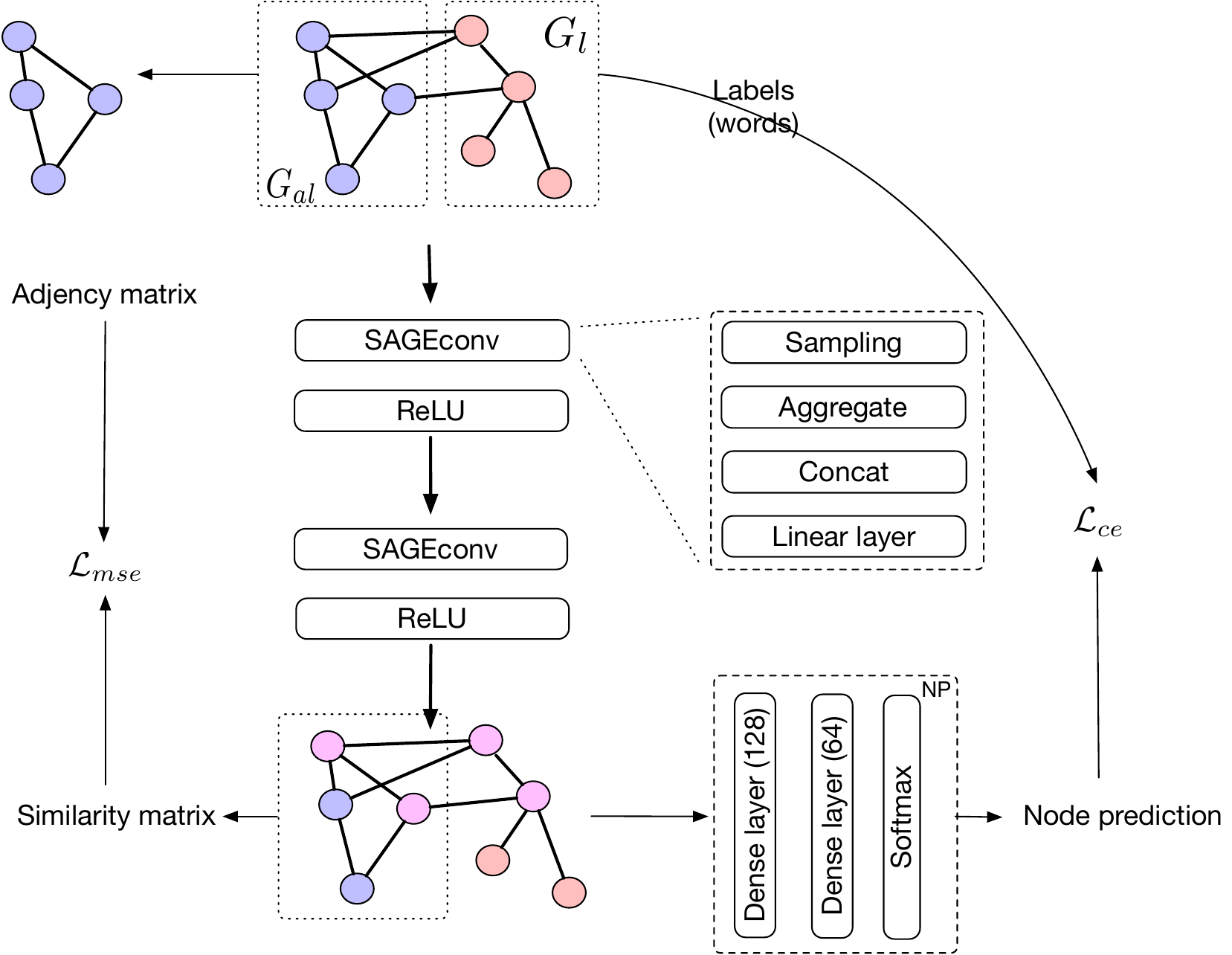}
         \caption{Training architecture of the graph convolutional neural network.}
        \label{fig:gcn_architecture}
    \end{subfigure}
    \hfill
    \begin{subfigure}{0.46\textwidth}
        \centering
        \includegraphics[width=\linewidth]{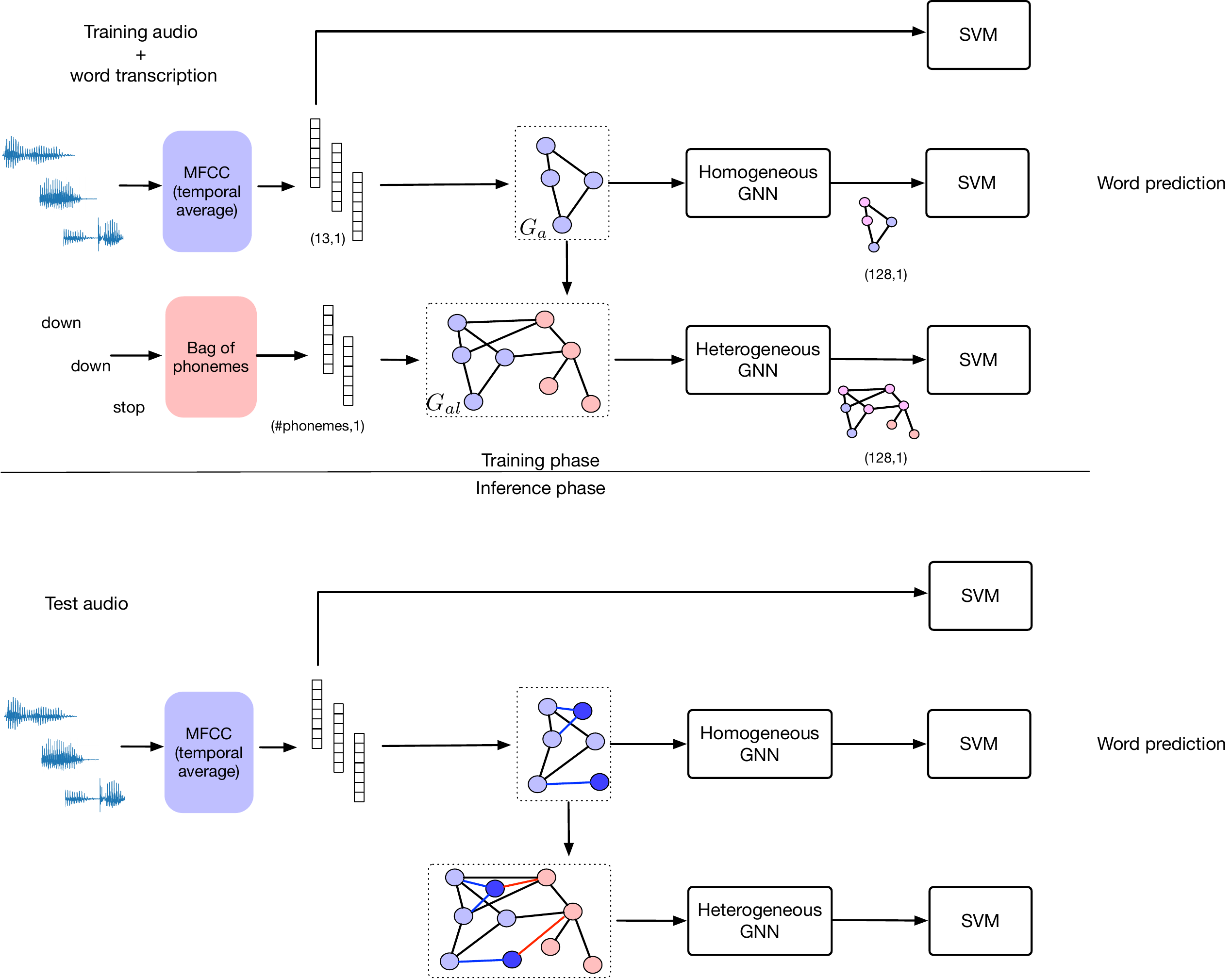}
       
        \caption{Experimental protocol: training phase (top) and inductive test phase (bottom).}
        \label{fig:exp_protocol}
    \end{subfigure}

    \caption{Overview of the experimental protocol and the GCN training architecture.}
    \label{fig:global_arch}
\end{figure*}

As we can see in Fig.~\ref{fig:gcn_architecture}, the objective function has 2 parts. The first part is dedicated to node prediction (NP) --\textit{e.g.} word prediction-- with a \textit{cross entropy} loss function ($\mathcal{L}_{ce}$) defined on the node label. It is defined as:
\begin{equation}
\mathcal{L}_{ce} = -\frac{1}{N} \sum_{i=1}^N \sum_{c=1}^C y_{i,c} \log(\hat{y}_{i,c})
\end{equation}
where \(N\) represents the total number of nodes in the graph, and \(C\) is the total number of classes corresponding to the number of unique words for each datasets. The term \(y_{i,c}\) is a one-hot encoded true label for node \(i\) belonging to class \(c\), while \(\hat{y}_{i,c}\) denotes the predicted probability of node \(i\) being assigned to class \(c\). To obtain the predicted probability, we use a small dense neural network designated by NP. 

The second part  (eq.~\ref{eq:rmse}) optimizes the edge weights with a \textit{mean square error}  ($\mathcal{L}_{mse}$) between the adjacency matrix of the acoustic graph ($A$) and the similarity matrix ($S$) based on the representations of these nodes.
\begin{equation}\label{eq:rmse}
\mathcal{L}_{mse} = \frac{1}{N^2} \sum_{i=1}^N \sum_{j=1}^N \left( A_{i,j} - S_{i,j} \right)^2
\end{equation}
where \(A_{i,j}\) is the element of the adjacency matrix that represents the edge weight between nodes \(i\) and \(j\), and \(S_{i,j}\) is the similarity score between nodes \(i\) and \(j\), based on their representations. Here, \(N\) refers to the total number of nodes in the graph.

 The total objective function is expressed as a linear combination of the two previous losses: $\mathcal{L} =  \mathcal{L}_{ce} +  \lambda\mathcal{L}_{mse}$       
where \(\lambda\) is a hyperparameter that balances the influence of the two loss terms.
 
  \subsubsection{Inference}
  \label{subsubsec:inf}
Our method is an inductive approach, which means that it generates the representations of audio samples that have not been seen during the learning phase. During the inference phase, \textit{i.e.} the embedding extraction phase, we transform new audio samples into nodes and add them to the heterogeneous graph in two steps. Let $v_{a_{i}}$ a new node and $k_{inf} \in \mathbb{N^{*}}$ an hyperparameter used to select, neigbors of $v_{a_{i}}$ in $G_{a}$: (i) $k_{inf}$ nearest neighbors are selected in the acoustic graph $G_{a}$, then an edge $\{v_{a_{i}},v_{a_{j}}\}$ is added with weight \(SIM_{inf}(v_{a_{i}},v_{a_{j}}) \). 
(ii) edges between $v_{a_{i}}$ and the linguistic subgraph are added as described in section \ref{subsec:hetero}.

\subsection{Experimental protocol}
To evaluate the benefit of cross-modal knowledge transfer on acoustic representations, we followed a two-step experimental protocol (Fig.\ref{fig:exp_protocol}). First, a heterogeneous graph was constructed from words-utterances and their transcriptions, and two GCN variants were trained: an acoustic-only (homogeneous) GCN and a cross-modal acoustic-linguistic (heterogeneous) GCN. Word-level acoustic embeddings were then extracted from each model and used to train an SVM classifier for isolated word recognition. The SVM was chosen for its efficiency and its ability to reflect the quality of the representation without modifying it. Second, we evaluated the representations both intrinsically and extrinsically.
Clustering quality of acoustic nodes was measured using intra- and inter-class inertia, complemented by average similarity metrics, denoted by \textit{same-sim} the average similarity of node representing the same word;  \textit{diff-sim} the average similarity of node representing different words; in order to capture both cluster- and node-level structure.
The extrinsic impact was assessed via classification accuracy on the isolated word recognition task. During inference, new acoustic nodes were added to the graphs, embeddings were extracted, and SVM performance was evaluated. This protocol enables quantifying how linguistic information explicitly transferred through the heterogeneous GCN enhances both the structural properties and the practical utility of acoustic representations in low-resource speech settings.

In our experiments, we use three different similarity measures:
$SIM_l$, the cosine similarity between two linguistic nodes;
$SIM_a$, defined as the exponential of the negative Dynamic Time Warping (DTW) distance~\cite{dtw}, which is used to find the neighborhood of an acoustic node; and
$SIM_{inf}$, defined as the exponential of the negative Kullback--Leibler (KL) divergence between the probability distributions output by the acoustic model (AM) (Section~\ref{subsubsec:inf}).

Our approach, as a proof of concept, is designed with rather simple choices. Further refinements will be needed on similarity measures and acoustic/linguistic embeddings to improve the performances.

\subsection{Datasets}
Two datasets of isolated speech recognition task are used. For each 80\% are used to train the AM and the HGNN, while the remaining 20\% are used in the inductive evaluation. 
\begin{enumerate}
    \item \textit{Google speech commands (GC)\cite{speechcommandsv2}} is a reduced version(to fit low ressources settings) of the standard Benchmark  Google speech command dataset, which contains 3,000 recordings 8 words in 16kHz performed by a wide variety of speakers. The number of word occurrences is uniformly distributed and covers 17 phonemes provided by the \textit{eng\_ipa} module of Python. 
    %
    \item \textit{Yemba dataset 
    (YD)\cite{KanaAzeuko2024}} consists of 1,753 recordings of 13 isolated words in Yemba language(one language from West Cameroon) extract from \cite{KanaAzeuko2024},  in 44.1kHz; where 180 speakers pronounce each word at least once. Th number of word occurances is nearly balanced, and covers 27 phonemes obtained by manual inventory. The words selected for experimentation are the simplest ones from the original dataset, which also contains groups of words.
\end{enumerate}


  


\subsection{Implementation}\label{sec:implementation}
For each audio recording, we constrain the size to a length of 16000. We extract 13 MFCCs coefficients using a frame length of 128 and a window size of 255. Parameters are consistent within a given dataset.
We implement two GCN models as described in Figure~\ref{fig:exp_protocol}; a homogeneous GCN with two layers of convolution (SARGEconv) for acoustic graph only and a heterogeneous GCN with two layers of convolution also where we have one convolutionnal block for each sub-graph (Acoustic graph, Linguistic graph, Acoustic-Linguistic graph). As mentioned in sec.~\ref{subsubsec:model} we use \textit{Mean aggregator}~\cite{GraphSAGE}. Both GCNs deliver an output of size 128. For the hyperparameters of the model we choose to fix $\delta_{a} = 0.0$ , $\
\delta_{l}=\delta_{al}=0.1$, and $\lambda=1$, then for rest the best values has been $k_{in}=0.5\times(\frac{size\,\, of \,\, dataset}{number\,\, of\,\, words} -1), k_{b}=\frac{k_{in}}{4},k_{inf}=\frac{k_{in}}{2}, \lambda=1$.

As mentioned in sec.\ref{subsec:hetero}, we independently pre-train the \(AM\) model, to produce a distributional probability over the target words to be recognized from all training data comming from the target dataset. The model is implemented as a simple dense neural network (DNN) with two hidden layers consisting of 128 and 64 neurons, respectively. The training is performed on the training subset of our dataset for each experiment. The \(AM\) model takes MFCC features as input and outputs a probability vector of size \((\text{number of words} \times 1)\), representing the likelihood of each word. We also use a DNN with 
a similar architecture to the NP block (see Figure~\ref{fig:gcn_architecture}) 
to get a cross-entropy loss relative to a node prediction. This dense neural network has 2 hidden layers of sizes respectively 64 and 32 with a dropout of 0.5 before the output layer. We trained and evaluated three SVMs using a linear kernel with default parameters
(sklearn 0.22.1) to isolate the impact of representation quality.

\section{Results and discussions}

The results highlight the intrinsic and extrinsic properties of the GNN representations. Extrinsic properties will be analyzed in terms of performance on the isolated word recognition task, while intrinsic properties will be analyzed in terms of the quality of the clusters formed from the representations. 

\subsection{Isolated words recognition results}
Table \ref{tab:combined} shows the accuracy of the SVM that uses MFCCs (denoted by ``MFCC'') considering as a baseline and GCN representations trained based on acoustic modality as input representations (denoted by ``ACOUS''), and cross-modal GCN representations trained on the basis of acoustic and linguistic modalities (denoted by ``CM''). We can observe that SVMs based on features from cross-modal GCN  outperform the baseline and the monomodal GCN on any dataset. This highlights the effectiveness and advantages of transferring linguistic knowledge to acoustic representation through heterogeneous graph neural networks. Furthermore, confidence intervals ($CI = \pm 1.96\sqrt{ac(1-ac)/N_b}$, where $ac$ denotes the accuracy and $N_b$ the number of samples used to compute $ac$~\cite{tahon:hal-01404146}), were computed on CM accuracy. These intervals confirm a significant performance difference with respect to the baseline MFCC and ACOUS approaches, whose accuracies fall well below the lower bound of the confidence interval, thus confirming the effectiveness of the transfer on GC. However, this is not observed for YD, probably due to the limited size of the test dataset.


\begin{table}[!h]
 
    \centering 
 \caption{Comparison of SVM accuracy (and 95\% confidence intervals) on different datasets.}
    \label{tab:combined}   
        \begin{tabular}{|c|c|c|c|}
            \midrule
            Dataset & MFCC & ACOUS  & CM \\ 
            \hline
            GC & 49.0 {\scriptsize$\pm 4.0$}  & 66.3 {\scriptsize$\pm 3.8$}& \textbf{72.3} {\scriptsize$\pm 3.5$}\\
            YD &  72.1 {\scriptsize$\pm 4.7$}  & 74.4 {\scriptsize$\pm 4.6$}  & \textbf{76.1} {\scriptsize$\pm  4.5$}  \\
            \bottomrule
        \end{tabular}
\end{table}

\begin{table}[!h]
\small
\centering
\caption{Comparison of clustering metrics across datasets for ACOUS and CM approaches.}
\label{tab:clustering}
\begin{tabular}{|l|cc|cc|}
\toprule
\multirow{2}{*}{Metric} 
& \multicolumn{2}{c|}{ACOUS} 
& \multicolumn{2}{c|}{CM} \\
\cline{2-5}
& GC & YD & GC & YD \\
\midrule
Inter-class  & 23.35 & 62.85 & 24.49 & 78.24 \\
Intra-class  & 226   & 91    & 24             & 28.26 \\
Same-sim     & 0.95  & 0.94  & 0.89           & 0.89 \\
Diff-sim     & 0.89  & 0.86  & 0.13           & 0.0612 \\
\bottomrule
\end{tabular}
\end{table}


\subsection{Clustering analysis}
As shown in Table \ref{tab:clustering}, the cross-modal approach exhibits higher inter-class inertia for all datasets, thus highlighting its ability to better separate different words. 
The intra-class inertia  significantly drop between both representations, favoring the cross modal approach, which achieves much lower intra-class inertia on GC and YD.

More specifically, for each dataset, the difference between the average cosine similarity for nodes representing the same target (``Same-sim'') and the average similarity for nodes with different targets (``Diff-sim'') is not nearly as pronounced as in the acoustic approach than in the cross-modal approach.
This suggests that the cross-modal GCN generates more distinct and word-specific representations compared to the acoustic GCN for each dataset. This observation emphases the transfer knowledge from linguistic graph to acoustic graph and his benefit on the acoustic representations which are enriched by incorporating linguistic information through the cross-modal GCN.

\section{Conclusion}
  \label{sec:con}

This work demonstrates that we can perform an explicit and beneficial cross-modal knowledge transfer from linguistic to acoustic representations through HGNN. 
This transfer leaded to consistent improvements in isolated word recognition across two datasets (Google Commands and Yemba) and intrinsic properties of embeddings (clustering) in low-resource settings. 
However further refinements are then needed to improve representation quality (similarity measures, acoustic/linguistic embeddings,  etc.) and move towards sentence level representations. 



\bibliographystyle{IEEEtran}
\bibliography{sn-bibliography}

\end{document}